\documentclass{ecai} 

\usepackage{latexsym}
\usepackage{amssymb}
\usepackage{amsmath}
\usepackage{amsthm}
\usepackage{booktabs}
\usepackage{enumitem}
\usepackage{graphicx}
\usepackage{cleveref}
\usepackage{bm}
\usepackage{tabularray}
\usepackage[acronym]{glossaries}
\usepackage{color}

\newcommand{\BibTeX}{B\kern-.05em{\sc i\kern-.025em b}\kern-.08em\TeX}

\newacronym{ai}{AI}{Artificial Intelligence}
\newacronym[plural=ANNs,firstplural=Artificial Neural Networks (ANNs)]{ann}{ANN}{Artificial Neural Network}

\newacronym{cfg}{CFG}{Context-free Grammar}
\newacronym[plural=CPUs,firstplural=Central Processing Units (CPUs)]{cpu}{CPU}{Central Processing Unit}
\newacronym[plural=CNNs,firstplural=Convolutional Neural Networks (CNNs)]{cnn}{CNN}{Convolutional Neural Network}

\newacronym{dl}{DL}{Deep Learning}
\newacronym[plural=DNNs,firstplural=Deep Neural Networks (DNNs)]{dnn}{DNN}{Deep Neural Network}
\newacronym{denser}{DENSER}{Deep Evolutionary Network Structured Evolution}
\newacronym{dsge}{DSGE}{Dynamic Structured Grammatical Evolution}

\newacronym{ea}{EA}{Evolutionary Algorithm}
\newacronym{ec}{EC}{Evolutionary Computation}
\newacronym[plural=ES,firstplural=Evolution Strategies (ES)]{es}{ES}{Evolution Strategy}

\newacronym[plural=LLMs,firstplural=Large Language Models (LLMs)]{llm}{LLM}{Large Language Model}
\newacronym{lemonade}{LEMONADE}{Lemarckian Evolutionary Algorithm for Multi-Objective Neural Architecture Design}

\newacronym{fdenser}{Fast-DENSER}{Fast Deep Evolutionary Network Structured Representation}
\newacronym{flops}{FLOPs}{floating point operations per second}

\newacronym{ge}{GE}{Grammatical Evolution}
\newacronym{gp}{GP}{Genetic Programming}

\newacronym{iot}{IoT}{Internet of Things}

\newacronym{ml}{ML}{Machine Learning}
\newacronym{moo}{MOO}{Multi-Objective Optimization}
\newacronym{mol}{MOL}{Multi-Output Learning}
\newacronym{monas}{MONAS}{Multi-Objective Neural Architecture Search}
\newacronym{mac}{MAC}{Multiply–accumulate operation}

\newacronym{nas}{NAS}{Neural Architecture Search}
\newacronym{ne}{NE}{Neuroevolution}
\newacronym{nsga-ii}{NSGA-II}{Nondominated Sorting Genetic Algorithm II}
\newacronym{nvml}{NVML}{NVIDIA Management Library}

\newacronym[plural=SNNs,firstplural=Spiking Neural Networks (SNNs)]{snn}{SNN}{Spiking Neural Network}

\newacronym{psu}{PSU}{Power Supply Unit}

\newacronym{rnn}{RNN}{Recurrent Neural Network}

\newacronym{sge}{SGE}{Structured Grammatical Evolution}

\newcommand{\rowsep}{2pt} 
\newcommand{\rowsepBigTables}{2.3pt} 

\begin{document}


\begin{frontmatter}


\paperid{} 

\title{GreenFactory: Ensembling Zero-Cost Proxies to Estimate Performance of Neural Networks}


\author[A]{\fnms{Gabriel}~\snm{Cortês}\thanks{Corresponding Author. Email: cortes@dei.uc.pt}}
\author[A]{\fnms{Nuno}~\snm{Lourenço}}
\author[B]{\fnms{Paolo}~\snm{Romano}} 
\author[A]{\fnms{Penousal}~\snm{Machado}} 

\address[A]{University of Coimbra, CISUC/LASI, DEI}
\address[B]{INESC-ID \& Instituto Superior Técnico, Universidade de Lisboa}


\begin{abstract}
Determining the performance of a Deep Neural Network during Neural Architecture Search processes is essential for identifying optimal architectures and hyperparameters. Traditionally, this process requires training and evaluation of each network, which is time-consuming and resource-intensive. Zero-cost proxies estimate performance without training, serving as an alternative to traditional training. However, recent proxies often lack generalization across diverse scenarios and provide only relative rankings rather than predicted accuracies. To address these limitations, we propose GreenFactory, an ensemble of zero-cost proxies that leverages a random forest regressor to combine multiple predictors' strengths and directly predict model test accuracy. We evaluate GreenFactory on NATS-Bench, achieving robust results across multiple datasets. Specifically, GreenFactory achieves high Kendall correlations on NATS-Bench-SSS, indicating substantial agreement between its predicted scores and actual performance: 0.907 for CIFAR-10, 0.945 for CIFAR-100, and 0.920 for ImageNet-16-120. Similarly, on NATS-Bench-TSS, we achieve correlations of 0.921 for CIFAR-10, 0.929 for CIFAR-100, and 0.908 for ImageNet-16-120, showcasing its reliability in both search spaces.
\end{abstract}

\end{frontmatter}


\section{Introduction}

The impact that \gls{ai} has had on human lives and society spans various domains, from advances in healthcare diagnosis \cite{Alowais2023} to optimization of trade routes \cite{Vaddy_2023}, identification of diseases in plants \cite{Alatawi2022PlantDD}, and cyber intrusion detection \cite{Dash2022ThreatsAO}, allowing us to improve most fields of study. However, these advancements come at a cost, with several issues affecting \gls{ai} systems: racial and gender bias in automated job and loan applications \cite{Kadiresan2022}, misclassification in critical scenarios due to data manipulation by bad actors \cite{RamirezAttacks}, job displacement \cite{UnemploymentAI,GomesIRobotAIEthics}, and massive energy consumption \cite{SustainableAIWynsberghe}. The latter is quantifiable and has grown exponentially in recent years. For instance, the popular GPT-3 model, similar to the models behind OpenAI's ChatGPT, used 1287 MWh in 15 days just for its training \cite{patterson2022carbon}. The current ``arms race'' to develop \glspl{llm} with more capabilities will likely increase the number of parameters, ensuring that newer models consume even more energy. Moreover, serving millions of users on a daily basis requires a tremendous quantity of processing devices, such as GPUs and/or TPUs, which use a massive amount of energy. The total energy consumption of NVIDIA GPUs, the leading manufacturer of these devices, is expected to surpass the total energy needs of countries such as Belgium or Switzerland \cite{DevriesEnergy,XuLLMSurvey}. Furthermore, technological companies like Google and Microsoft, which have previously committed to offsetting their carbon emissions, have struggled to maintain these promises due to allocating more resources to \gls{ai} data centers \cite{MSNMicrosoftCarbonEmissions,Google2024EnvReport}. This growth is expected to continue, with hardly any restrictions, due to this technology's positive impact on the population and the economy \cite{AIEconomy}.

Several techniques have been proposed to address the energy consumption problem of \gls{ai}, particularly of \gls{ml}. For instance, using \gls{nas} or \gls{ne} to search for energy-efficient \gls{dnn} models has also significantly increased energy efficiency \cite{CortesEvoAPPS24}. These approaches, however, require substantial resources since each \gls{dnn} must undergo a full training process and subsequent evaluation, which may ultimately undermine its potential for minimal energy usage. Zero-cost proxies have emerged as a promising solution to mitigate this. These techniques allow the estimation of network performance using only a fraction of the traditionally required computational resources \cite{TrainingFreeNASReview}. However, they can only be used to rank different models instead of directly estimating their final quality. Additionally, most proxies struggle to generalize across different datasets. Some perform well on specific datasets but poorly on others. Combining the strengths of multiple proxies, therefore, holds great potential to achieve more robust and generalized performance estimations.

This paper proposes GreenFactory, a novel technique that integrates a diverse array of zero-cost proxies from the literature using a random forest regression model. Our approach begins by calculating the scores produced by these proxies on the NATS-Bench benchmark, covering both size and topology search spaces. This information is aggregated into a unified dataset with the corresponding search space and dataset associated with each proxy score. We employ a 70-15-15 split for training, validation, and testing to identify the most suitable regression model for this dataset. A recursive feature elimination process is then conducted to reduce the number of features, effectively minimizing the number of proxies required by the model. This process identifies two optimal configurations: one achieving the best root mean squared error (RMSE) and another offering a balanced trade-off between RMSE and proxy computing time. In this context, RMSE measures how closely the model's predicted performance aligns with the actual performance of architectures in NATS-Bench. We develop the latter to reduce the inherent cost of computing multiple zero-cost proxies to obtain the ensemble model's prediction. The hyperparameters of both configurations are fine-tuned through standard hyperparameter optimization and subsequently evaluated on the NATS-Bench across all search spaces and datasets. Additionally, the evaluation incorporates a stratified sampling approach to address the skewness in the benchmark, ensuring a more balanced and robust assessment.

Our main contribution is developing a robust model capable of accurately predicting the test accuracy of a \gls{dnn}. This represents an advancement over the current state-of-the-art, where existing zero-cost proxies are limited to generating network scores without directly estimating accuracy. Experimentally, on the NATS-Bench-SSS, our model achieved Kendall correlations of 0.907 for CIFAR-10, 0.945 for CIFAR-100, and 0.920 for ImageNet-16-120. Similarly, for the NATS-Bench-TSS, we obtained correlations of 0.921 for CIFAR-10, 0.929 for CIFAR-100, and 0.908 for ImageNet-16-120.

The structure of this paper is as follows: \Cref{sec:background} provides an overview of the essential background on zero-cost proxies and the benchmark used. Next, \Cref{sec:methodology} describes the workings of the proposed algorithm and details the experimental setup. \Cref{sec:results} presents the results obtained, accompanied by an in-depth discussion. Finally, \Cref{sec:conclusion} summarizes the key findings and suggests potential avenues for future research.


\section{Background and Related Work}
\label{sec:background}
\subsection{Zero-Cost Proxies}
Zero-cost proxies, also known as the training-free method, are techniques that allow the prediction of the quality of a model without training it \cite{TrainingFreeNASReview}. This is achieved through algorithms or mathematical formulas that estimate how good a model might be. Zero-cost proxies have been a relatively recent research thread in the \gls{ml} community since they were introduced in 2018 \cite{CameroLowCostExpectedPerformance} and have seen continuous improvement. Traditional \gls{nas} methods typically require hundreds or thousands of GPU hours and, as such, can significantly benefit from using training-free methods, which usually require only a small amount of GPU time or even CPU time. Despite numerous statistical analyses of training-free NAS algorithms, a theoretical analysis of training-free algorithms remains lacking. A comprehensive theoretical examination of this score function is essential to further research in this field \cite{TrainingFreeNASReview}. To avoid training many networks to evaluate the correlation between the proxy and the actual accuracy, well-established \gls{nas} datasets are used \cite{TrainingFreeNASReview}. Among other metrics, these datasets contain the architectures of many networks and their corresponding accuracies.

A pioneer in the field, \textbf{NASWOT} is an algorithm that generates scores reflecting a model’s test accuracy without requiring training. It computes these scores based on the network’s activation patterns in response to a single mini-batch \cite{MellorNASWOT}. \textbf{Synflow} is a pruning algorithm that aims to prevent the layer collapse problem when pruning a neural network. It was extended as a data-independent estimator of a network's performance without training it \cite{Synflow}. Inspired by pruning-at-initialization, \textbf{GradNorm} is a proxy metric that computes the sum of the Euclidean norms of the gradients after passing a single mini-batch of training data \cite{Gradnorm}. \textbf{TE-NAS} evaluates network architectures by examining the spectrum of the neural tangent kernel and the number of linear regions in the input space \cite{TENAS}. \textbf{Zen-NAS} is a zero-shot method that measures the expressivity of a \gls{dnn} by computing its Zen-Score, which is derived from a few forward inferences on randomly initialized networks with random Gaussian inputs \cite{ZenNAS}. \textbf{ZiCo} demonstrates that high-performance \glspl{dnn} tend to possess high absolute mean values and low standard deviation values for the gradient and uses that information to estimate the performance of a network \cite{Zico}.

\textbf{EZNAS} employs a \gls{gp} approach to automate the discovery of zero-cost proxies for \gls{nas} scoring, achieving results comparable to state-of-the-art. However, its evaluation lacks clarity regarding other proxies, relies on partial layer statistics, and uses a low recombination rate, potentially limiting consistent optimization. Like EZNAS, \textbf{GreenMachine} is an evolutionary approach for discovering zero-cost proxies, demonstrating competitive performance in correlation with the actual test accuracy of the \gls{dnn} models \cite{greenmachine}. Additionally, GreenMachine uses a stratified dataset sampling, ensuring the proxies effectively distinguish between low- and high-performing models. This strategy minimizes overfitting and improves the generalizability of the discovered proxies.

\textbf{AZ-NAS} introduces four complementary zero-cost proxies that assess expressivity, progressivity, trainability, and complexity (FLOPs) in architectures. The proxy scores are computed in a single forward and backward pass. AZ-NAS uses a non-linear ranking aggregation method to identify consistently high-ranked networks across proxies. Evaluated on 3,000 networks, AZ-NAS achieves state-of-the-art results in multiple search spaces, demonstrating its effectiveness in identifying high-quality architectures. \textbf{Expressivity} measures a network's ability to represent diverse semantics by analyzing feature space isotropy with random weights and Gaussian inputs. Higher isotropy implies less feature correlation, greater semantic storage capacity, and reduced risk of dead neurons or dimensional collapse. The \textbf{progressivity} proxy measures a network's ability to expand its feature space with depth, indicating its potential to capture high-level semantics. It is quantified as the smallest difference in expressivity scores between consecutive blocks, with high scores reflecting steady growth in expressivity. The \textbf{trainability} proxy assesses how easily gradients can propagate through a network during initialization, which is crucial for training. It analyzes gradient behavior within individual primary blocks, employing an efficient approximation of the Jacobian matrix to reduce computational overhead. This method approximates the Jacobian matrix by feeding a Rademacher random vector into the block's backward pass and computing the resulting input gradients. The spectral norm of the approximate Jacobian serves as the stability measure. 

\subsection{NATS-Bench}
Benchmarking \gls{nas} algorithms is challenging due to variations in data preprocessing, evaluation pipelines, and random seeds. Moreover, fully training many networks requires extensive computational resources, thus making reproducibility difficult for most researchers. To address this issue, some datasets were proposed to standardize the benchmarking process, providing a common ground for comparisons and reducing the required computational resources by delivering the results that would otherwise require the complete training of many \gls{dnn} models. These benchmarks typically include not only the network configuration and its accuracy metric after training but also other data such as latency, number of parameters, and more, enabling a quicker assessment of the proposed \gls{nas} method.

NAS-Bench-201 features full graph cells, allowing for a more comprehensive search space, though limited to four nodes and five associated operation options to maintain manageability \cite{NASBench201}. Each architecture is trained and evaluated on the CIFAR-10, CIFAR-100, and ImageNet-16-120 image classification datasets. It contains 15,625 architectures. NATS-Bench is a unified benchmark for searching for architecture topology and size \cite{NATSBench}. It includes 15,625 candidates for the architecture topology (TSS) and 32,768 for architecture size (SSS). It also presents results for CIFAR-10, CIFAR-100, and ImageNet-16-120.


\section{GreenFactory}
\label{sec:methodology}

\paragraph{Data Collection}

We use the NATS-Bench benchmark, focusing on its two search spaces: size (NATS-Bench-SSS) and topology (NATS-Bench-TSS). The zero-cost proxies incorporated into the ensemble are NASWOT, Synflow, GradNorm, TE-NAS, Zen-NAS, ZiCo, EZNAS, AZ-NAS, expressivity, progressivity, trainability, and the 10 most promising proxies obtained by the GreenMachine algorithm, complemented by the number of trainable parameters and FLOPs of each network. The GreenMachine proxies include the three described in \cite{greenmachine}, and their formulas are presented as supplementary material. Note that the individual performance of these proxies is less critical since the ensemble model determines their importance, assigning less weight to poorly performing proxies.  

We compute the scores for these proxies on the NATS-Bench-SSS and NATS-Bench-TSS for the CIFAR-10, CIFAR-100, and ImageNet-16-120 datasets. To process the GreenMachine proxies more efficiently, we compute them for each benchmark instance, avoiding redundant steps such as feature extraction and associated mini-batch data processing, as network features remain constant across multiple proxy evaluations for the same network.  

\begin{figure}[h]
    \centering
    \includegraphics[width=\linewidth]{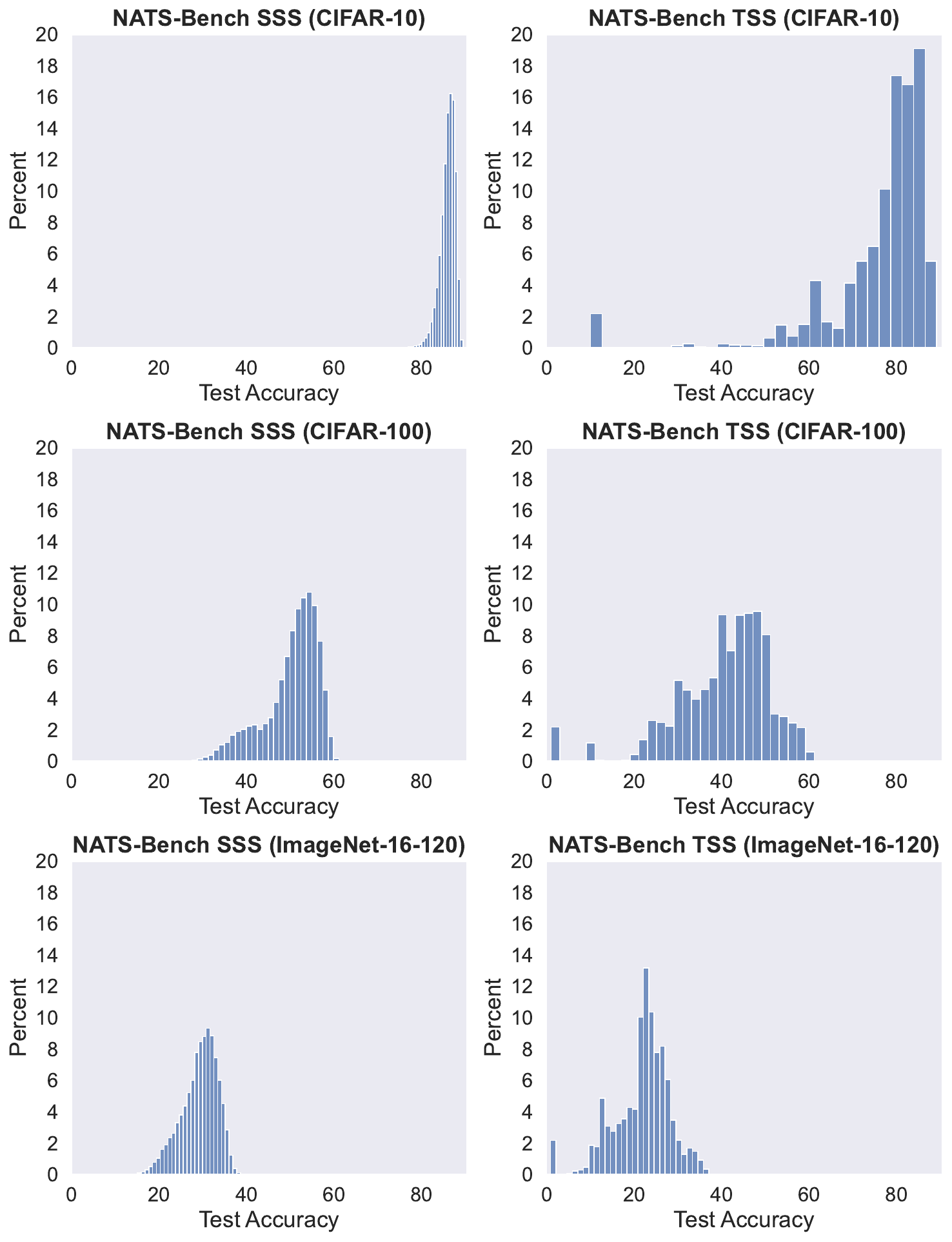}
    \caption{Histograms of test accuracy for the NATS-Bench benchmark across the topology and size search spaces and the CIFAR-10, CIFAR-100, and ImageNet-16-120 datasets.}
   \label{fig:nats_bench_histogram}
\end{figure}

The resulting dataset, \emph{Green-NATS-Bench}, consists of the scores obtained by 21 zero-cost proxies, the number of trainable parameters, the number of FLOPs, and the target variable, which is the test accuracy for each of the CIFAR-10, CIFAR-100, and ImageNet-16-120 datasets. The dataset also includes each network's corresponding search space and dataset information, represented with one-hot encoding.

\Cref{fig:nats_bench_histogram} illustrates the distribution of test accuracy across the two search spaces and three datasets. A skew towards higher accuracy values is evident, likely due to the networks achieving better performance after extensive training epochs and the tendency of larger networks in NATS-Bench-SSS to correlate with higher accuracy.  

The dataset is divided into training, validation, and testing sets, with the test set accounting for 15\% of the total data, the validation set comprising 15\% as well, and the remaining 70\% used for training. In line with the methodology proposed in \cite{greenmachine}, we also create a stratified version of the dataset, dividing it into five bins based on test accuracy. This stratification is intended to ensure that the ensemble proxy accurately predicts the performance of high-performing networks and identifies and discards poorly performing ones. Additionally, we developed another version of the dataset where all features are standardized.  

\paragraph{Regression Model Selection}
As an exploratory step, we experiment with multiple regression algorithms and assess their performance based on the RMSE. To avoid introducing bias, we exclude the search space information from the dataset during this step while retaining details about the image classification dataset. This mirrors feeding a batch of actual data to the network to obtain layer features when computing zero-cost proxies. Additionally, the significant variations in ranges of test accuracy values across the three datasets challenge the ensemble model's performance. Since the goal is to estimate test accuracy directly rather than a relative score, these discrepancies in accuracy ranges can hinder estimation, as the model must reconcile differing scales and distributions. We conduct this process for 30 random seeds and report the mean RMSE and the standard deviation.  

\paragraph{Feature Selection}
We begin by incorporating all the selected zero-cost proxies into the ensemble proxy. However, increasing the number of proxies also increases computation time, as each proxy must be calculated. Since the ensemble aims to achieve high accuracy and time efficiency, we perform a feature selection step using feature ranking with recursive feature elimination (RFE) \cite{RFE}. RFE is an iterative method that ranks features based on their importance and removes the least important ones, reducing the dimensionality of the feature set. This process helps to refine the model by retaining the most relevant features while minimizing computational cost, ensuring the ensemble maintains its predictive power.

In addition to evaluating the RMSE, we also assess the computational efficiency by measuring the time required to calculate the proxies for each feature subset. To accomplish this, we randomly sample 100 networks from each combination of search space and dataset, recording the total time taken to compute all of them. It is important to note that, akin to the data collection process, the GreenMachine proxies and the four AZ-NAS proxies are calculated simultaneously, leveraging implementation optimizations to improve performance.

To balance performance and computational efficiency, we assign a score that reflects the trade-off between the two factors. This score is the mean of the normalized RMSE and the normalized measured time. We select the model with the lowest RMSE (indicating better performance) and the model that achieves the best score (a lower score indicates a better balance of accuracy and time efficiency).  

\paragraph{Hyperparameter Tuning}
We use Optuna \cite{optuna_2019} for hyperparameter optimization, using the Tree-structured Parzen Estimator \cite{bergstra_tpe} sampler to efficiently explore the search space and minimize the objective function. The optimization process comprises 1,000 trials. The objective metric for each trial is the RMSE obtained by the random forest regression model on the validation subset, using the hyperparameters suggested by the sampler.

Hyperparameter tuning is performed for the best model in terms of RMSE and for the model that obtains the optimal balance between error and time efficiency. The set of allowed hyperparameter values in the model optimization is presented in \Cref{tab:optuna_params}.

\begin{table}[h]
\centering
\caption{Set of allowed hyperparameters in the hyperparameter optimization.}
\label{tab:optuna_params}
\begin{tblr}{
  hline{1,8} = {-}{0.08em},
  hline{2} = {-}{},
  rowsep = \rowsep,
}
Parameter         & Values                                       \\
Estimators        & $[10, 1000]$                                 \\
Max features      & $[1, 10] \cup \{\text{sqrt}, \text{log2}\}$  \\
Min samples split & $[2, 32]$                                    \\
Min samples leaf  & $[1, 32]$                                    \\
Max depth         & $[10, 100]$                                  \\
Bootstrap         & $\{\text{False}, \text{True}\}$                        
\end{tblr}
\end{table}

\paragraph{Evaluation}
The state-of-the-art zero-cost proxies and our ensemble models are evaluated on the test set of \emph{Green-NATS-Bench} across two search spaces and three datasets. This test set represents 15\% of the overall dataset, including 4,915 networks from the NATS-Bench-SSS search space and 1,552 networks from the NATS-Bench-TSS. This evaluation encompasses a significantly larger number of benchmark networks compared to the original methodologies employed by most existing proxies, enabling a more comprehensive study of their quality.

The proxies are compared using Kendall's and Spearman's rank correlation coefficients, providing a robust measure of how well they align with the actual test accuracy \cite{TrainingFreeNASReview}. Additionally, we report the RMSE of our ensemble models across the search spaces and datasets, offering a detailed evaluation of their predictive performance. This analysis is not feasible for the other proxies, as they predict only relative rankings rather than accuracy values.

\paragraph{Experimental Setup}
Data collection was conducted on a machine with Ubuntu 22.04.3 LTS, two Intel Xeon Silver 4310 CPUs (2.10GHz, 12 cores each), 256 GB RAM, and three NVIDIA RTX A6000 GPUs (48 GB GDDR6 each), using CUDA 12.1, Python 3.10, and PyTorch 2.3.1. Additional experiments were run on a Fujitsu PRIMERGY CX2550 M5 with two Intel Xeon Gold 6148 CPUs (2.40GHz), 384 GB DDR4 RAM, Python 3.10, Scikit-learn 1.6.0, and Optuna 4.1.0.


\section{Results and Discussion}
\label{sec:results}
\paragraph{Regression Model Selection}
\Cref{tab:regression_types_results} presents the regression model selection process results in terms of mean RMSE over 30 runs. Results for the SGD regressor are shown only for the standardized dataset, as this regressor is highly sensitive to feature scaling and performs poorly on unstandardized data. The random forest regressor consistently outperforms other models across all datasets, achieving the lowest RMSE overall in the stratified dataset. The linear regressor and Bayesian ridge perform poorly, especially on datasets other than the standardized one. The linear regressor’s simple linear assumption fails to capture more complex, non-linear relationships in the data, and its sensitivity to outliers and inability to handle feature interactions contribute to its higher error. Similarly, the Bayesian ridge model's assumptions may not be flexible enough to model more intricate patterns. 

\begin{table}[h]
\centering
\caption{Comparison of regressor performance across different sampling strategies in terms of mean RMSE over 30 runs. The best value for each strategy is highlighted in bold.}
\label{tab:regression_types_results}
\begin{tblr}{
  width = \linewidth,
  colspec = {Q[200]Q[250]Q[250]Q[244]},
  hline{1,11} = {-}{0.08em},
  hline{2} = {-}{},
  rowsep = \rowsep
}
Regressor          & None                 & Stratified           & Standardized         \\
Bayesian Ridge     & $24.62 \pm 0.67$     & $25.26 \pm 0.06$     & $4.08 \pm 0.02$      \\
Linear             & $24.29 \pm 0.04$     & $25.26 \pm 0.06$     & $4.08 \pm 0.02$      \\
Decision Tree      & $1.56 \pm 0.02$      & $1.10 \pm 0.02$      & $1.69 \pm 0.02$      \\
Gradient Boosting  & $1.95 \pm 0.01$      & $2.13 \pm 0.02$      & $2.04 \pm 0.02$      \\
Random Forest      & \bm{$1.05 \pm 0.01$} & \bm{$0.76 \pm 0.02$} & \bm{$1.15 \pm 0.01$} \\
XGB                & $1.19 \pm 0.01$      & $1.12 \pm 0.01$      & $1.28 \pm 0.02$      \\
ElasticNet         & $10.70 \pm 0.04$     & $10.71 \pm 0.04$     & $6.76 \pm 0.03$      \\
SGD                & --                   & --                   & $4.14 \pm 0.03$      \\
CatBoost           & $1.11 \pm 0.01$      & $1.07 \pm 0.01$      & $1.21 \pm 0.01$
\end{tblr}
\end{table}

Models like gradient boosting, decision trees, and XGB show moderate performance, with their results varying depending on the dataset's preprocessing. ElasticNet also shows high RMSE values, particularly when using non-standardized data. Based on these results, we select the random forest regression model in combination with the stratified dataset, as this pairing yielded the lowest RMSE across all models and dataset preprocessing types. A random forest model excels in this because it combines many decision trees and selects informative features through splits, allowing the capture of complex non-linear relationships in the data. Moreover, it is less sensitive to outliers and noise than other models.

\begin{figure}[h]
    \centering
    \includegraphics[width=\linewidth]{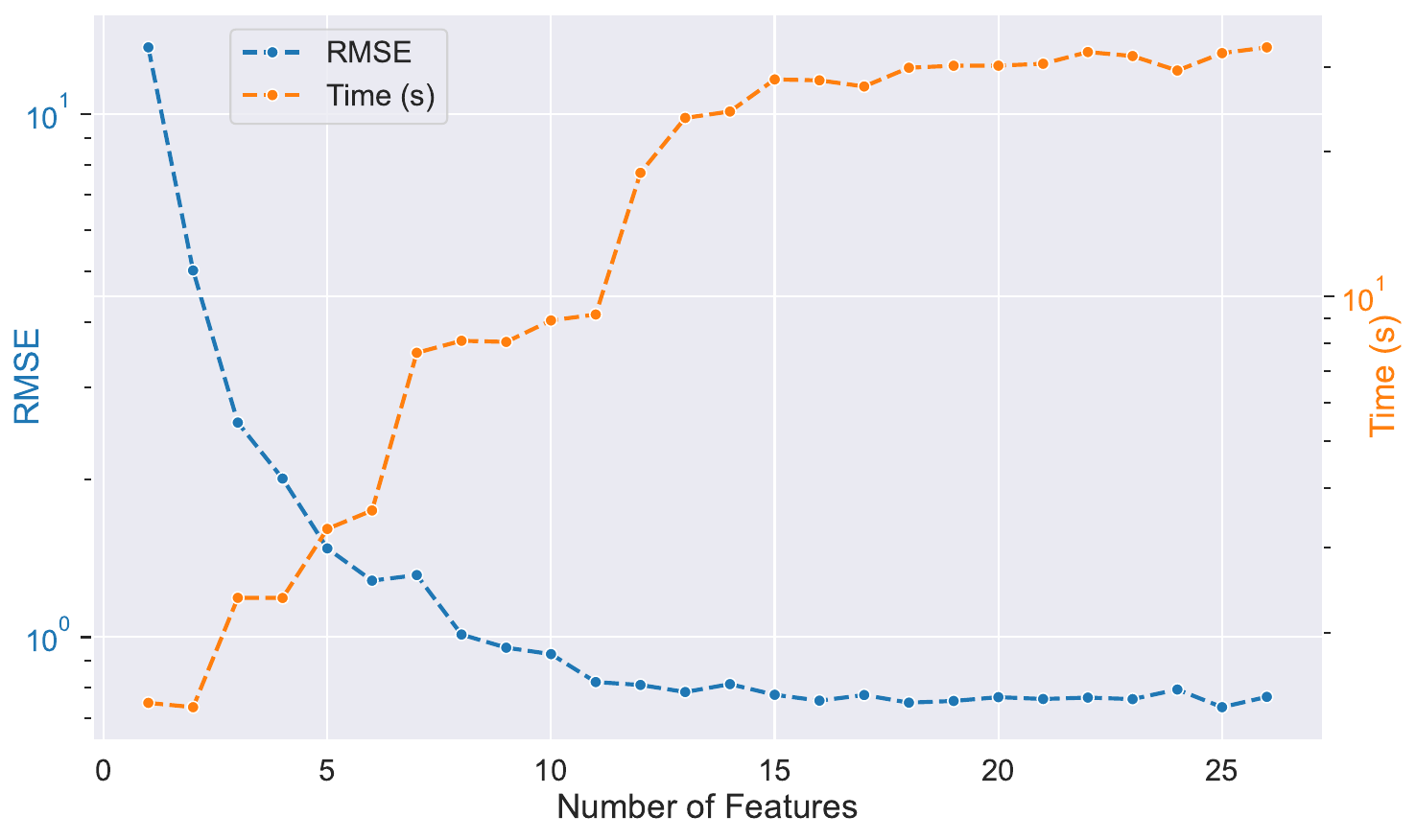}
    \caption{Relationship between the number of features and RMSE and mean computation time required to obtain the features for a single \gls{dnn}.}
   \label{fig:rfe_results}
\end{figure}

\paragraph{Feature Selection}
The results of the feature ranking with recursive feature elimination process are presented in \Cref{fig:rfe_results}, showcasing the evolution of the RMSE and of the duration required to compute the number of features, i.e., the zero-cost proxies, as the number of selected features is increased. Detailed data from this analysis, such as the calculated scores obtained using the previously described method and the selected features for each step, are provided in the supplementary material. The results are unsurprising: reducing the number of selected features generally degrades predictive performance, as the ensemble model loses the ability to capture data complexity. Conversely, increasing the number of features tends to improve performance moderately but substantially raises computational time, since most proxies demand relatively significant processing. Exceptions to this pattern include simple metrics like the number of parameters, FLOPs, and dataset indicators, which offer low computational overhead. Notably, there are points where additional features bring little to no RMSE improvement while still increasing cost, which highlights the importance of a balanced selection.

Considering these results, we select the case where 25 features are chosen, as it achieves the lowest RMSE, reducing it from 0.768 to 0.734 compared to using all features. Notably, this selection excludes only the Synflow proxy. However, it is crucial to address the computational cost of obtaining features for each network, as this approach is expected to significantly reduce the time required compared to training the networks directly. To balance predictive performance and computational efficiency, we opt for a reduced set of six features, achieving an RMSE of 1.282 while requiring only 11\% of the computation time compared to the scenario with the lowest RMSE. This optimized feature set includes three GreenMachine proxies (E, F, and J), GradNorm, EZ-NAS, and the CIFAR-10 dataset indicator. The selection of CIFAR-10 over other dataset indicators is justified by its distinct test accuracy distribution across both search spaces (\Cref{fig:nats_bench_histogram}). While this configuration may not maximize predictive performance, it is adequate given the constraints on the number of features.

\paragraph{Hyperparameter Tuning}
The best hyperparameters found in the optimization process are presented in \Cref{tab:optuna_optimized_params} for the GreenFactory (GF) and GreenFactory-Fast (GF-Fast) models, with GF-Fast utilizing fewer features for improved efficiency. Their performance in terms of RMSE over the two search spaces and three datasets is detailed in \Cref{table:optimized_model_results}. 

As expected, the GF model attains lower errors in all considered cases since it uses about 4 times more features than its counterpart, albeit requiring more computational time. In both models, RMSE values are generally higher for the NATS-Bench-TSS search spaces compared to the NATS-Bench-SSS. This difference reflects the increased complexity of NATS-Bench-TSS, as it encompasses networks that vary in topology, whereas the other focuses only on network size.

\begin{table}
\centering
\caption{Hyperparameters found by the hyperparameter optimization for the GreenFactory and GreenFactory-Fast models.}
\label{tab:optuna_optimized_params}
\begin{tblr}{
  hline{1,8} = {-}{0.08em},
  hline{2} = {-}{},
  rowsep = \rowsep,
}
Parameter         & GreenFactory & GreenFactory-Fast \\
Estimators        & 968          & 945               \\
Max features      & 8            & 10                \\
Min samples split & 3            & 3                 \\
Min samples leaf  & 1            & 1                 \\
Max depth         & 38           & 73                \\
Bootstrap         & False        & False             
\end{tblr}
\end{table}

\begin{table}[h]
\centering
\caption{Performance of the regression model after hyperparameter optimization (GreenFactory) and of the one that uses fewer proxies after hyperparameter optimization (GreenFactory-Fast). The results reported as RMSE are provided for the CIFAR-10, CIFAR-100, and ImageNet-16-120 datasets across the two search spaces.}
\label{table:optimized_model_results}
\begin{tblr}{
  row{2} = {c},
  row{9} = {c},
  cell{2}{1} = {c=3}{},
  cell{9}{1} = {c=3}{},
  hline{1-2,16} = {-}{},
  hline{3,9-10} = {-}{dashed},
  rowsep = \rowsep,
}
Dataset               & RMSE (GF) & RMSE (GF-Fast) \\
Randomly sampled data &           &                \\
SSS (CF-10)           & 0.273     & 0.621          \\
SSS (CF-100)          & 0.721     & 1.780          \\
SSS (IN-16-120)       & 0.562     & 1.440          \\
TSS (CF-10)           & 1.071     & 2.508          \\
TSS (CF-100)          & 1.251     & 3.493          \\
TSS (IN-16-120)       & 0.920     & 2.368          \\
Stratified data       &           &                \\
SSS (CF-10)           & 0.274     & 0.664          \\
SSS (CF-100)          & 0.616     & 1.454          \\
SSS (IN-16-120)       & 0.515     & 1.372          \\
TSS (CF-10)           & 1.044     & 2.479          \\
TSS (CF-100)          & 0.979     & 3.013          \\
TSS (IN-16-120)       & 0.546     & 1.333          
\end{tblr}
\end{table}

\begin{figure}[h]
    \centering
    \includegraphics[width=\linewidth]{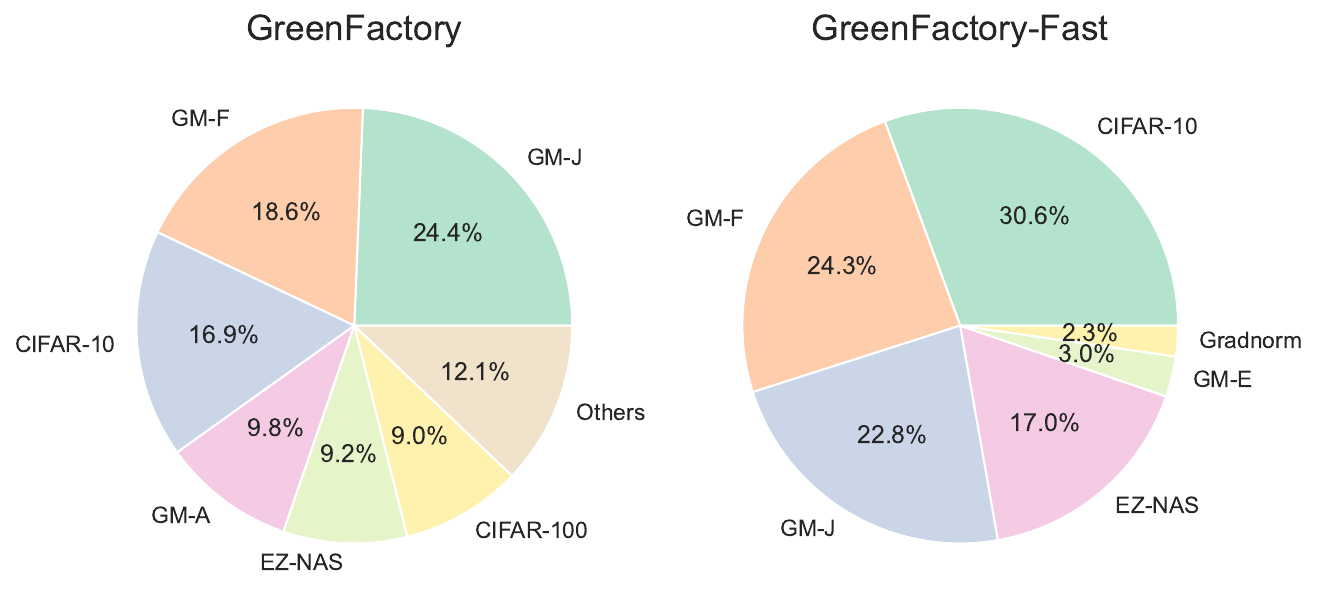}
    \caption{Feature importance on the two optimized models: GreenFactory and GreenFactory-Fast. For clarity, features with importance below 3\% are grouped under the category \emph{Others} in the GreenFactory model.}
   \label{fig:feature_importances}
\end{figure}

As previously discussed, the regression model assigns relative importance to each feature, making the individual performance of specific proxies less critical. Moreover, these feature importances are on the basis of the recursive feature elimination process used to reduce the number of selected features. \Cref{fig:feature_importances} provides a visual representation comparing the importance assigned to each feature in the optimized GreenFactory and GreenFactory-Fast models after training. For clarity, features with importance below 3\% are grouped under the category \emph{Others} in the case of the GreenFactory model. In both models, GreenMachine proxies contribute significantly to overall feature importance. When combined with EZ-NAS, they account for an even larger portion of the importance, highlighting the importance of zero-cost proxies developed through evolutionary methods.

\paragraph{Evaluation}

\Cref{table:results_random,table:results_stratified} provide a detailed comparison of the Kendall and Spearman correlation coefficients achieved by state-of-the-art proxies and our ensemble models, measuring the correlation between the computed scores and the test accuracy of the networks after training. The performance of the remaining zero-cost proxies (GreenMachine-A through G and AZ-NAS expressivity, progressivity, and trainability) are detailed as supplementary material.

The GreenFactory model represents the configuration that consistently achieves the lowest RMSE, excelling across all search spaces and datasets for both metrics. In contrast, GreenFactory-Fast is designed to balance performance and computational efficiency, consistently achieving runner-up results. Both models underwent a hyperparameter tuning process to optimize their performance.

Moreover, our solutions present consistent performance across both tested scenarios: when networks are randomly sampled from the dataset and when they are stratified. This highlights the robustness of our approach in predicting the performance of both low- and high-performing \gls{dnn} models.

\begin{table*}[hp]
\centering
\caption{Comparison of Zero-Cost proxies on the NATS-Bench benchmark across the CIFAR-10, CIFAR-100, and ImageNet-16-120 datasets on the test set of \emph{Green-NATS-Bench}, with \textbf{randomly sampled data}. The reported values represent the \textbf{absolute Kendall ($\tau$) and Spearman ($\rho$) correlation coefficients} measured on the test set of \emph{Green-NATS-Bench}. The best-performing value for each case is highlighted in bold.}
\label{table:results_random}
\begin{tblr}{
  cell{1}{2} = {c=2}{},
  cell{1}{4} = {c=2}{},
  cell{1}{6} = {c=2}{},
  cell{1}{8} = {c=2}{},
  cell{1}{10} = {c=2}{},
  cell{1}{12} = {c=2}{},
  hline{1,18} = {-}{0.08em},
  hline{3,16} = {-}{},
  rowsep = \rowsepBigTables,
  row{16} = {font=\bfseries},
  colspec = {Q[l] Q[c] Q[c] Q[c] Q[c] Q[c] Q[c] Q[c] Q[c] Q[c] Q[c] Q[c] Q[c]},
}
Proxy                 & SSS (CF-10) &   & SSS (CF-100) &   & SSS (IN-16-120) &   & TSS (CF-10) &   & TSS (CF-100) &   & TSS (IN-16-120) &  &  \\
                      & $\tau$ & $\rho$ & $\tau$ & $\rho$  & $\tau$ & $\rho$    & $\tau$ & $\rho$ & $\tau$ & $\rho$  & $\tau$ & $\rho$     &  \\
\#Params	          & 0.665& 0.852& 0.523& 0.705& 0.650& 0.828& 0.377& 0.529& 0.357& 0.497& 0.290& 0.405 \\
FLOPs                 & 0.404& 0.562& 0.181& 0.264& 0.407& 0.561& 0.377& 0.529& 0.357& 0.497& 0.290& 0.405 \\
Synflow               & 0.765& 0.926& 0.586& 0.781& 0.788& 0.931& 0.385& 0.546& 0.357& 0.514& 0.359& 0.511 \\
Gradnorm              & 0.180& 0.269& 0.495& 0.686& 0.380& 0.544& 0.143& 0.205& 0.033& 0.045& 0.022& 0.031 \\
NASWOT                & 0.378& 0.537& 0.164& 0.243& 0.391& 0.548& 0.434& 0.605& 0.403& 0.568& 0.391& 0.550 \\
TE-NAS                & 0.342& 0.490& 0.269& 0.390& 0.365& 0.516& 0.234& 0.308& 0.183& 0.247& 0.094& 0.125 \\
Zen-NAS               & 0.727& 0.900& 0.463& 0.644& 0.659& 0.836& 0.182& 0.264& 0.185& 0.266& 0.206& 0.295 \\
ZiCo                  & 0.734& 0.905& 0.560& 0.757& 0.728& 0.898& 0.384& 0.543& 0.351& 0.508& 0.344& 0.494 \\
EZNAS                 & 0.749& 0.917& 0.514& 0.699& 0.567& 0.752& 0.526& 0.709& 0.443& 0.615& 0.403& 0.570 \\
AZ-NAS                & 0.301& 0.430& 0.195& 0.289& 0.429& 0.588& 0.550& 0.741& 0.457& 0.640& 0.378& 0.537 \\
GreenMachine-1	      & 0.514& 0.706& 0.563& 0.732& 0.519& 0.714& 0.479& 0.652& 0.480& 0.655& 0.407& 0.572 \\
GreenMachine-2	      & 0.690& 0.869& 0.629& 0.814& 0.782& 0.933& 0.322& 0.459& 0.315& 0.453& 0.308& 0.441 \\
GreenMachine-3	      & 0.755& 0.916& 0.666& 0.843& 0.701& 0.877& 0.055& 0.090& 0.072& 0.113& 0.265& 0.374 \\
GreenFactory 	      & 0.907  & 0.986  & 0.945  & 0.994   & 0.920  & 0.990     & 0.921  & 0.987  & 0.929  & 0.989   & 0.908  & 0.980         \\
GreenFactory-Fast     & 0.790  & 0.930  & 0.876  & 0.969   & 0.804  & 0.937     & 0.823  & 0.935  & 0.809  & 0.918   & 0.760  & 0.884
\end{tblr}
\end{table*}

\begin{table*}[hp]
\centering
\caption{Comparison of Zero-Cost proxies on the NATS-Bench benchmark across the CIFAR-10, CIFAR-100, and ImageNet-16-120 datasets on the test set of \emph{Green-NATS-Bench}, with the data \textbf{stratified} by test accuracy, as described. The reported values represent the \textbf{absolute Kendall ($\tau$) and Spearman ($\rho$) correlation coefficients} measured on the test set of \emph{Green-NATS-Bench}. The best-performing value for each case is highlighted in bold.}
\label{table:results_stratified}
\begin{tblr}{
  cell{1}{2} = {c=2}{},
  cell{1}{4} = {c=2}{},
  cell{1}{6} = {c=2}{},
  cell{1}{8} = {c=2}{},
  cell{1}{10} = {c=2}{},
  cell{1}{12} = {c=2}{},
  row{16} = {font=\bfseries},
  hline{1,18} = {-}{0.08em},
  hline{3,16} = {-}{},
  rowsep = \rowsepBigTables,
  colspec = {Q[l] Q[c] Q[c] Q[c] Q[c] Q[c] Q[c] Q[c] Q[c] Q[c] Q[c] Q[c] Q[c]},
}
Proxy                 & SSS (CF-10)  &  & SSS (CF-100) &   & SSS (IN-16-120) &   & TSS (CF-10) &   & TSS (CF-100) &   & TSS (IN-16-120) &    &  \\
                      & $\tau$ & $\rho$ & $\tau$ & $\rho$  & $\tau$ & $\rho$    & $\tau$ & $\rho$ & $\tau$ & $\rho$  & $\tau$ & $\rho$     &  \\
\#Params	          & 0.661& 0.849& 0.572& 0.764& 0.706& 0.874& 0.402& 0.551& 0.412& 0.568& 0.317& 0.441 \\
FLOPs                 & 0.402& 0.558& 0.196& 0.286& 0.499& 0.674& 0.402& 0.551& 0.412& 0.568& 0.317& 0.441 \\
Synflow	              & 0.763& 0.926& 0.601& 0.799& 0.828& 0.954& 0.557& 0.740& 0.521& 0.704& 0.453& 0.633 \\
Gradnorm	          & 0.172& 0.257& 0.431& 0.603& 0.436& 0.621& 0.305& 0.433& 0.165& 0.248& 0.145& 0.215 \\
NASWOT	              & 0.377& 0.535& 0.174& 0.257& 0.478& 0.657& 0.622& 0.806& 0.530& 0.709& 0.512& 0.689 \\
TE-NAS	              & 0.347& 0.497& 0.286& 0.412& 0.428& 0.600& 0.132& 0.140& 0.009& 0.053& 0.090& 0.119 \\
Zen-NAS	              & 0.722& 0.898& 0.481& 0.670& 0.722& 0.887& 0.319& 0.459& 0.403& 0.562& 0.356& 0.521 \\
ZiCo	              & 0.730& 0.903& 0.564& 0.764& 0.778& 0.930& 0.570& 0.751& 0.503& 0.682& 0.463& 0.652 \\
EZNAS	              & 0.750& 0.917& 0.554& 0.750& 0.632& 0.814& 0.662& 0.848& 0.556& 0.737& 0.416& 0.583 \\
AZ-NAS                & 0.265& 0.381& 0.210& 0.312& 0.498& 0.664& 0.653& 0.848& 0.528& 0.719& 0.480& 0.663 \\
GreenMachine-1        & 0.511& 0.701& 0.537& 0.687& 0.561& 0.758& 0.607& 0.791& 0.624& 0.799& 0.448& 0.613 \\
GreenMachine-2        & 0.691& 0.870& 0.618& 0.806& 0.815& 0.950& 0.496& 0.669& 0.456& 0.626& 0.415& 0.575 \\
GreenMachine-3        & 0.750& 0.912& 0.700& 0.878& 0.744& 0.910& 0.233& 0.354& 0.248& 0.375& 0.437& 0.614 \\
GreenFactory          & 0.905  & 0.985  & 0.949  & 0.994   & 0.934  & 0.992     & 0.948  & 0.994  & 0.952  & 0.995   & 0.982  &  0.998        \\
GreenFactory-Fast     & 0.793  & 0.931  & 0.891  & 0.972   & 0.834  & 0.951     & 0.897  & 0.974  & 0.865  & 0.953   & 0.958  &  0.987
\end{tblr}
\end{table*}


\section{Conclusion}
\label{sec:conclusion}
This paper presents novel techniques to harness the strengths of multiple zero-cost proxies through an ensemble regression approach. We explore various regression models, conduct feature selection, and fine-tune hyperparameters to enhance predictive performance. Additionally, we develop a model that balances performance and computational efficiency. These strategies culminate in the GreenFactory and GreenFactory-Fast models.

The models are evaluated on the NATS-Bench benchmark using CIFAR-10, CIFAR-100, and ImageNet-16-120 datasets, demonstrating superior correlation metrics compared to the state-of-the-art across all scenarios. Specifically, GreenFactory achieves Kendall correlations of 0.907 on CIFAR-10, 0.945 on CIFAR-100, and 0.920 on ImageNet-16-120 with the NATS-Bench-SSS benchmark. Similarly, GreenFactory attains Kendall correlations of 0.921 on CIFAR-10, 0.929 on CIFAR-100, and 0.908 on ImageNet-16-120 with the NATS-Bench-TSS benchmark, highlighting its robust performance across diverse scenarios. Moreover, these models directly predict test accuracy, surpassing the relative performance scores typically provided by zero-cost proxies in the literature.

\subsection{Future Work}
While comprehensive, NATS-Bench is limited in dataset and architecture diversity. Expanding to additional benchmarks could enhance the generalizability of our solutions. Additionally, incorporating more zero-cost proxies and network characteristics could improve our ensemble models' robustness and predictive performance across various scenarios. Future work could explore replacing one-hot encoding with abstract features that describe the dataset (e.g., input/output size, task type, and number of classes). A leave-one-out test using NATS-Bench datasets could also assess ensemble generalization to unseen datasets.



\begin{ack}
By using the \texttt{ack} environment to insert your (optional) 
acknowledgements, you can ensure that the text is suppressed whenever 
you use the \texttt{doubleblind} option. In the final version, 
acknowledgements may be included on the extra page intended for references.
\end{ack}



\bibliography{mybibfile}

\clearpage
\newpage

\paragraph{Supplementary Material}

\begin{figure}[h]
    \centering
    \includegraphics[width=\linewidth]{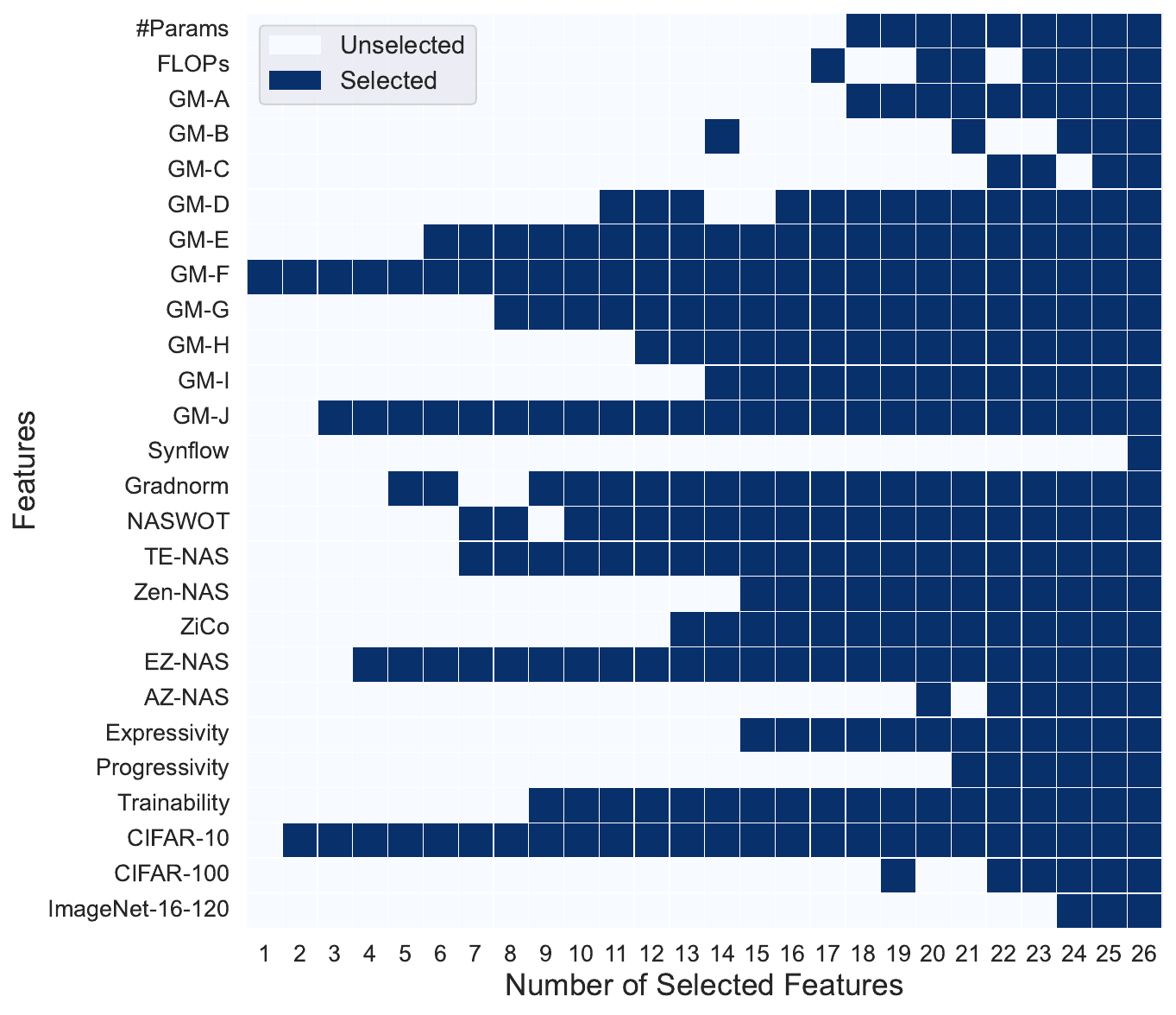}
    \caption{Features selected in the feature selection step.}
   \label{fig:rfe_heatmap}
\end{figure}

\Cref{table:rfe} presents the detailed results of the recursive feature elimination process, including RMSE, time to compute the selected proxies, and the score that allows us to assess the balance between performance and time efficiency. \Cref{fig:rfe_heatmap} presents the features selected for each step of the feature selection process. Note that the features \emph{CIFAR-10}, \emph{CIFAR-100}, and \emph{ImageNet-16-120} serve to indicate the dataset.

\Cref{fig:greenmachine} presents the formulas of the selected GreenMachine zero-cost proxies used in the paper. Note that the last three (Proxies H, I, J) are the ones described in the GreenMachine paper.

\Cref{table:resultsnon_stratified_kendall,table:resultsstratified_kendall} present a comparison of the zero-cost proxies that were not listed in the paper.

\begin{table}[h]
\centering
\caption{Results of the recursive feature elimination process. Bold denotes the best obtained value.}
\label{table:rfe}
\begin{tblr}{
  hline{1,28} = {-}{0.08em},
  hline{2} = {-}{},
  rowsep = \rowsep
}
Features & RMSE     & Time (s) & Score  \\
1        & 13.411   &  1.429   & 0.500  \\
2        &  5.022   &  \textbf{1.399}   & 0.169  \\
3        &  2.571   &  2.363   & 0.088  \\
4        &  2.009   &  2.361   & 0.066  \\
5        &  1.477   &  3.286   & 0.059  \\
6        &  1.282   &  3.589   & \textbf{0.056}  \\
7        &  1.314   &  7.637   & 0.122  \\
8        &  1.011   &  8.089   & 0.117  \\
9        &  0.954   &  8.044   & 0.114  \\
10       &  0.928   &  8.916   & 0.127  \\
11       &  0.820   &  9.173   & 0.127  \\
12       &  0.810   & 18.075   & 0.267  \\
13       &  0.785   & 23.509   & 0.352  \\
14       &  0.813   & 24.243   & 0.365  \\
15       &  0.776   & 28.264   & 0.427  \\
16       &  0.756   & 28.143   & 0.425  \\
17       &  0.775   & 27.332   & 0.412  \\
18       &  0.749   & 29.893   & 0.452  \\
19       &  0.755   & 30.180   & 0.457  \\
20       &  0.767   & 30.195   & 0.458  \\
21       &  0.761   & 30.472   & 0.462  \\
22       &  0.766   & 32.230   & 0.490  \\
23       &  0.761   & 31.618   & 0.480  \\
24       &  0.794   & 29.494   & 0.447  \\
25       &  \textbf{0.734}   & 32.059   & 0.486  \\
26       &  0.768   & 32.956   & 0.501
\end{tblr}
\end{table}

\begin{figure*}[h]
    \small
    {
    \begin{verbatim}
GreenMachine-A:
    frobenius_norm(greater_than(less_than(pass_perturbation_fwd_input,
    (greater_than((less_than(random_grad, abs(pass_perturbation_fwd_input))),
    (min(pass_noise_bwd_output, abs((min(subtract(pass_perturbation_fwd_input,
    (cosine_similarity(pass_perturbation_wt, pass_noise_bwd_output))),
    relu(pass_noise_fwd_input))))))))), pass_noise_fwd_output))

GreenMachine-B:
    (sum((max((equal(pass_perturbation_bwd_input, pass_perturbation_grad)),
    softmax(pass_noise_fwd_input))), random_grad))

GreenMachine-C:
    frobenius_norm((sum(cosine_similarity(less_than_zero(greater_than(power(element_wise_product((
    cosine_similarity(pass_noise_bwd_input,pass_noise_fwd_input)), pass_perturbation_bwd_input)),
    numel(subtract((subtract(pass_noise_fwd_output, sum(power(min(pass_perturbation_grad,
    random_grad)), determinant((element_wise_product(random_wt, subtract(pass_perturbation_grad,
    (equal(pass_perturbation_bwd_input, pass_noise_bwd_output))))))))), pass_perturbation_wt)))),
    random_grad), equal(pass_perturbation_fwd_output, pass_noise_bwd_input))))

GreenMachine-D: 
    softmax(softmax(pass_grad))

GreenMachine-E: 
    kl_div(l1_norm(pass_fwd_output), pass_perturbation_wt)

GreenMachine-F: 
    frobenius_norm(greater_than(pass_noise_bwd_input, pass_noise_fwd_output))

GreenMachine-G: 
    frobenius_norm(min(pass_wt, log(pass_noise_grad)))
    
GreenMachine-H (GreenMachine-3):
    cosine_similarity(softmax(cosine_similarity(pass_perturbation_bwd_output, pass_fwd_output)),
    (greater_than(greater_than(pass_noise_fwd_output, cosine_similarity(pass_perturbation_grad,
    less_than(less_than(equal((equal(abs((max(transpose(pass_noise_bwd_output),
    normalize(less_than_zero(pass_fwd_output))))),
    (greater_than(gaussian_init((kl_div(power(pass_noise_bwd_input), normalized_sum(pass_grad)))),
    element_wise_product(kl_div(cosine_similarity(pass_perturbation_fwd_output, random_grad),
    (sum(random_wt, pass_perturbation_grad))), greater_than(pass_bwd_output,
    (max(pass_perturbation_fwd_output, pass_bwd_input)))))))),
    (subtract(element_wise_invert(pass_perturbation_wt),
    gaussian_init(determinant(pass_perturbation_bwd_input))))),
    (greater_than(kl_div((min(random_grad, pass_noise_bwd_input)), pass_noise_fwd_output),
    (mat_mul((element_wise_product(pass_noise_fwd_input, pass_perturbation_fwd_output)),
    ones_like(sum((mat_mul(frobenius_norm(pass_fwd_input),
    frobenius_norm(pass_perturbation_bwd_output))), numel(l1_norm(pass_noise_grad))))))))),
    greater_than(pass_noise_fwd_input, (subtract(sum(random_wt, heaviside(pass_noise_bwd_output)),
    softmax((mat_mul(sum(greater_than(pass_fwd_output,
    cosine_similarity(pass_perturbation_bwd_output, pass_perturbation_fwd_input)),
    (sum((min(pass_bwd_output, pass_noise_wt)), cosine_similarity(pass_grad, pass_noise_grad)))),
    sigmoid(frobenius_norm(normalize(pass_fwd_output)))))))))))), random_wt)))

GreenMachine-I (GreenMachine-2): 
    max(softmax(pass_perturbation_fwd_output),
    equal(pass_perturbation_grad, element_wise_product(pass_perturbation_bwd_input,
    pass_bwd_input)))

GreenMachine-J (GreenMachine-1): 
    (greater_than((mat_mul(pass_noise_wt,
    (greater_than((kl_div(pass_noise_grad, pass_noise_wt)), subtract(pass_noise_fwd_output,
    pass_perturbation_bwd_input))))), random_wt))
    \end{verbatim}
    }
    \caption{Formulas of the selected GreenMachine zero-cost proxies.}
    \label{fig:greenmachine}
\end{figure*}

\begin{table*}[h]
\centering
\caption{Comparison of the remaining Zero-Cost proxies on the NATS-Bench benchmark across the CIFAR-10, CIFAR-100, and ImageNet-16-120 datasets on the test set of \emph{Green-NATS-Bench}, with \textbf{randomly sampled data}. The reported values represent the \textbf{absolute Kendall ($\tau$) and Spearman ($\rho$) correlation coefficients} measured on the test set of \emph{Green-NATS-Bench}. The best-performing value for each case is highlighted in bold.}
\label{table:resultsnon_stratified_kendall}
\begin{tblr}{
  cell{1}{2} = {c=2}{},
  cell{1}{4} = {c=2}{},
  cell{1}{6} = {c=2}{},
  cell{1}{8} = {c=2}{},
  cell{1}{10} = {c=2}{},
  cell{1}{12} = {c=2}{},
  hline{1,13} = {-}{0.08em},
  hline{3} = {-}{},
  rowsep = \rowsepBigTables,
  colspec = {Q[l] Q[c] Q[c] Q[c] Q[c] Q[c] Q[c] Q[c] Q[c] Q[c] Q[c] Q[c] Q[c]},
}
Proxy                 & SSS (CF-10) &   & SSS (CF-100) &   & SSS (IN-16-120) &   & TSS (CF-10) &   & TSS (CF-100) &   & TSS (IN-16-120) &  &  \\
                      & $\tau$ & $\rho$ & $\tau$ & $\rho$  & $\tau$ & $\rho$    & $\tau$ & $\rho$ & $\tau$ & $\rho$  & $\tau$ & $\rho$   &  \\
GreenMachine-A        & 0.557& 0.751& 0.338& 0.488& 0.566& 0.750& 0.237& 0.352& 0.210& 0.309& 0.202& 0.295 \\
GreenMachine-B        & 0.712& 0.887& 0.576& 0.765& 0.746& 0.913& 0.364& 0.488& 0.356& 0.481& 0.322& 0.447 \\
GreenMachine-C        & 0.095& 0.138& 0.160& 0.234& 0.251& 0.369& 0.418& 0.548& 0.394& 0.520& 0.377& 0.494 \\
GreenMachine-D        & 0.714& 0.888& 0.578& 0.766& 0.750& 0.914& 0.393& 0.557& 0.374& 0.532& 0.345& 0.493 \\
GreenMachine-E        & 0.653& 0.841& 0.587& 0.782& 0.704& 0.881& 0.370& 0.517& 0.340& 0.487& 0.329& 0.468 \\
GreenMachine-F        & 0.551& 0.744& 0.331& 0.479& 0.559& 0.743& 0.357& 0.510& 0.321& 0.458& 0.283& 0.409 \\
GreenMachine-G        & 0.693& 0.868& 0.374& 0.515& 0.523& 0.674& 0.092& 0.099& 0.093& 0.087& 0.123& 0.152 \\
Expressivity          & 0.175& 0.179& 0.192& 0.247& 0.533& 0.672& 0.458& 0.638& 0.365& 0.510& 0.252& 0.368 \\
Progressivity         & 0.075& 0.104& 0.061& 0.089& 0.094& 0.145& 0.226& 0.331& 0.233& 0.338& 0.189& 0.285 \\
Trainability          & 0.194& 0.282& 0.073& 0.109& 0.107& 0.158& 0.429& 0.592& 0.353& 0.495& 0.387& 0.555 \\
\end{tblr}
\end{table*}

\begin{table*}[hp]
\centering
\caption{Comparison of the remaining Zero-Cost proxies on the NATS-Bench benchmark across the CIFAR-10, CIFAR-100, and ImageNet-16-120 datasets on the test set of \emph{Green-NATS-Bench}, with the data \textbf{stratified} by test accuracy, as described. The reported values represent the \textbf{absolute Kendall ($\tau$) and Spearman ($\rho$) correlation coefficients} measured on the test set of \emph{Green-NATS-Bench}. The best-performing value for each case is highlighted in bold.}
\label{table:resultsstratified_kendall}
\begin{tblr}{
  cell{1}{2} = {c=2}{},
  cell{1}{4} = {c=2}{},
  cell{1}{6} = {c=2}{},
  cell{1}{8} = {c=2}{},
  cell{1}{10} = {c=2}{},
  cell{1}{12} = {c=2}{},
  hline{1,13} = {-}{0.08em},
  hline{3} = {-}{},
  colspec = {Q[l] Q[c] Q[c] Q[c] Q[c] Q[c] Q[c] Q[c] Q[c] Q[c] Q[c] Q[c] Q[c]},
}
Proxy                 & SSS (CF-10) &   & SSS (CF-100) &   & SSS (IN-16-120) &   & TSS (CF-10) &   & TSS (CF-100) &   & TSS (IN-16-120) &  &  \\
                      & $\tau$ & $\rho$ & $\tau$ & $\rho$  & $\tau$ & $\rho$    & $\tau$ & $\rho$ & $\tau$ & $\rho$  & $\tau$ & $\rho$   &  \\
GreenMachine-A        & 0.556& 0.750& 0.355& 0.513& 0.634& 0.815& 0.497& 0.674& 0.371& 0.527& 0.381& 0.530 \\
GreenMachine-B        & 0.714& 0.890& 0.573& 0.764& 0.788& 0.937& 0.540& 0.697& 0.482& 0.645& 0.444& 0.596 \\
GreenMachine-C        & 0.066& 0.094& 0.159& 0.232& 0.331& 0.478& 0.602& 0.748& 0.526& 0.673& 0.489& 0.629 \\
GreenMachine-D        & 0.717& 0.892& 0.576& 0.767& 0.790& 0.938& 0.486& 0.671& 0.435& 0.619& 0.400& 0.570 \\
GreenMachine-E        & 0.648& 0.838& 0.614& 0.812& 0.749& 0.911& 0.480& 0.646& 0.531& 0.706& 0.393& 0.548 \\
GreenMachine-F        & 0.549& 0.742& 0.348& 0.503& 0.628& 0.810& 0.544& 0.728& 0.463& 0.635& 0.397& 0.561 \\
GreenMachine-G        & 0.692& 0.868& 0.437& 0.601& 0.603& 0.764& 0.180& 0.254& 0.219& 0.275& 0.155& 0.223 \\
Expressivity          & 0.151& 0.139& 0.199& 0.257& 0.581& 0.707& 0.558& 0.754& 0.464& 0.635& 0.329& 0.468 \\
Progressivity         & 0.093& 0.132& 0.079& 0.114& 0.058& 0.096& 0.405& 0.579& 0.290& 0.438& 0.363& 0.512 \\
Trainability          & 0.178& 0.261& 0.094& 0.142& 0.103& 0.149& 0.477& 0.632& 0.467& 0.638& 0.416& 0.580 \\
\end{tblr}
\end{table*}

\end{document}